\documentclass[10pt,twocolumn,letterpaper]{article}

\usepackage{cvpr}              % To produce the CAMERA-READY version
\usepackage{tabularx}
\newcolumntype{Y}{>{\centering\arraybackslash}X}

\usepackage{bm}
\usepackage{multirow}
\usepackage{pifont}
\usepackage{makecell}
\usepackage[accsupp]{axessibility} % Improves PDF readability for those with disabilities.
\newcommand{\cmark}{\ding{51}} 
\newcommand{\xmark}{\ding{55}}

\definecolor{cvprblue}{rgb}{0.21,0.49,0.74}
\usepackage[pagebackref,breaklinks,colorlinks,allcolors=cvprblue]{hyperref}

\def\paperID{46176}
\def\confName{CVPR}
\def\confYear{2026}

\title{CoCoVideo: The High-Quality \underline{Co}mmercial-Model-Based \underline{Co}ntrastive Benchmark for AI-Generated Video Detection}

\author{
\begin{tabular}{c}
Huidong Feng$^{1}$,
Wentao Chen$^{2}$,
Jie Chen$^{2}$,
Xinqi Cai$^{1, 3}$,
Ruolong Ma$^{2}$,\\
Yinglin Zheng$^{1}$,
Yuxin Lin$^{1}$,
Ming Zeng$^{1}$\thanks{Corresponding author.} \\
{\small $^{1}$School of Informatics, Xiamen University} \\
{\small $^{2}$China Academy of Information and Communications Technology} \quad
{\small $^{3}$AI Transcend Pte. Ltd.}\\
{\tt\small \{fenghuidong@stu., caixinqi@stu., zhengyinglin@stu., linyx@stu., zengming@\}xmu.edu.cn} \\
{\tt\small \{chenwentao, chenjie7, maruolong\}@caict.ac.cn}
\end{tabular}
}

\begin{document}
\maketitle
\begin{abstract}
\vspace{2.0pt}

With the rapid advancement of artificial intelligence generated content (AIGC) technologies, video forgery has become increasingly prevalent, posing new challenges to public discourse and societal security. Despite remarkable progress in existing deepfake detection methods, AIGC forgery detection remains challenging, as existing datasets mainly rely on open-source video generation models with quality far below that of commercial AIGC systems.
Even datasets containing a few commercial samples often retain visible watermarks, compromising authenticity and hindering model generalization to high-fidelity AIGC videos.
To address these issues, we introduce \textbf{CoCoVideo-26K}, a contrastive, commercial-model-based AIGC video dataset covering 13 mainstream commercial generators and providing semantically aligned real–fake video pairs.
This dataset enables deeper exploration of the differences between authentic and high-quality synthetic videos and establishes a new benchmark for highly realistic video forgery detection.
Building on this dataset, we propose \textbf{CoCoDetect}, a detection framework integrating contrastive learning with confidence-gated multimodal large language model (MLLM) inference.
An R3D-18 backbone extracts spatio-temporal representations, while a confidence gate routes uncertain cases to an MLLM for reasoning about physical plausibility and scene consistency.
Extensive experiments on CoCoVideo-26K and public benchmarks demonstrate state-of-the-art performance, validating the framework’s robustness and generalizability.
Our code and dataset are available at \url{https://github.com/DonoToT/CoCoVideo}.
\end{abstract} 
\section{Introduction}
\label{sec:intro}
In recent years, generative adversarial networks (GANs)~\cite{goodfellow2014generative, nirkin2019fsgan} have driven significant progress in image and video synthesis.
Early deepfake technologies~\cite{deepfake2019, faceswap2018, thies2016face2face} leveraged convolutional neural networks (CNNs)~\cite{lecun2002gradient, bayar2016deep} and some graphics-based methods to generate face-swap videos.
While such techniques have been applied in media production, they have also been misused for identity fraud and other threats to public security~\cite{twomey2023deepfake, busacca2023deepfake}.
To mitigate these threats, early detection research primarily focused on identifying manipulation cues, such as visual artifacts~\cite{rossler2019faceforensics++, zhao2021multi, nguyen2024laa}, pose inconsistencies~\cite{guo2025face, nguyen2025vulnerability, zhang2024learning}, and frequency-domain abnormalities~\cite{kashiani2025freqdebias, kim2025beyond, tan2024frequency, wang2023dynamic}, to distinguish manipulated content from authentic media.

\begin{figure}[t]
  \centering
   \includegraphics[width=\columnwidth]{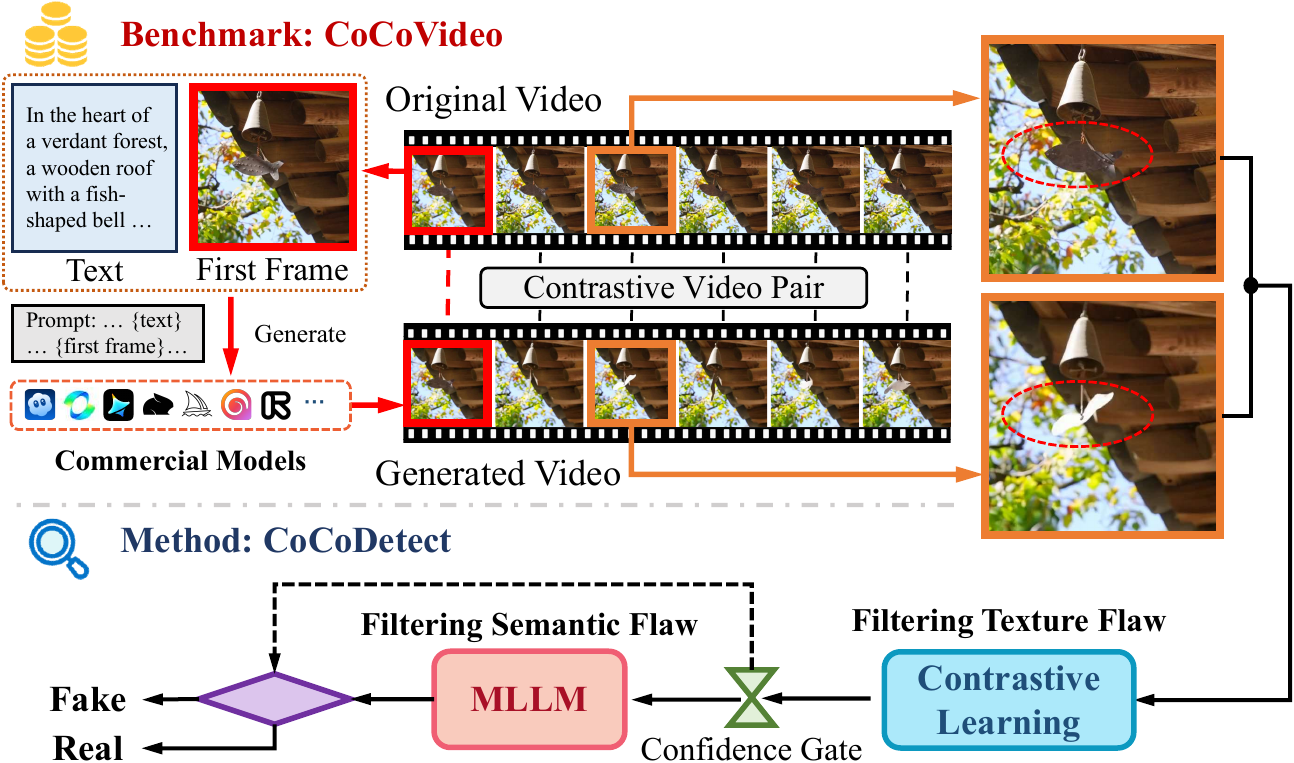}
   \caption{Overview of CoCoVideo dataset and CoCoDetect framework. CoCoVideo contains contrastive video pairs from 13 commercial models sharing identical first frames. CoCoDetect integrates dual-head contrastive learning with confidence-gated MLLM reasoning for AIGC video detection.}
   \label{fig:teaser}
   \vspace{-1.5em}
\end{figure}

At the same time, the rise of artificial intelligence generated content (AIGC) has reshaped the landscape of video synthesis: video generation has evolved from early GANs to text-to-video and image-to-video systems based on diffusion models~\cite{sohl2015deep, ho2020denoising}, achieving higher fidelity and more diverse outputs.
Traditional detection methods struggle to keep up with this new paradigm.
\textbf{First}, they are typically designed for specific generative architectures or manipulation operations, limiting their generalization ability and preventing effective use of high-level semantic cues, such as physical logic and scene coherence.
\textbf{Second}, the current research landscape lacks a high-quality and widely accessible AIGC video dataset.
Existing datasets, such as GenBuster~\cite{wen2025busterx} and GenVidBench~\cite{ni2025genvidbench}, are mainly constructed using open-source models~\cite{hong2022cogvideo, khachatryan2023text2video, hacohen2024ltx, kong2024hunyuanvideo}, whose outputs often exhibit low visual realism and lack convincing motion and scene consistency.
Models trained on such data tend to overfit these low-quality generative patterns and fail to generalize to high-fidelity commercial AIGC videos, ultimately limiting their real-world applicability.

To address these gaps, we first build a high-quality AIGC video forgery detection benchmark dataset, \textbf{CoCoVideo-26K}.
We utilize OpenVid-1M~\cite{nan2024openvid} as the real video source and generate forged videos mainly through high-quality \textit{commercial} video generation models.
This dataset is organized in a \textit{contrastive} format, where each real video is paired with a synthetic counterpart, forming one-to-one "real--fake" video pairs.
This structure provides stable positive and negative supervision for contrastive learning and facilitates fine-grained analysis of real-synthetic differences.
Compared with existing datasets that rely primarily on open-source models, the synthetic videos in CoCoVideo achieve substantially higher visual realism and stronger scene-level semantic coherence, enabling more reliable evaluation and training under real-world AIGC detection scenarios.

Then, based on CoCoVideo, we propose \textbf{CoCoDetect}, which integrates \textit{contrastive} learning with a \textit{confidence-gated} multimodal large language models (MLLMs) reasoning mechanism for AIGC video detection.
High-quality commercial AIGC videos often exhibit subtle and diverse texture-level patterns, making low-level artifact cues unreliable.
The tightly semantic-aligned real--fake pairs in CoCoVideo enable the model to learn fine-grained appearance differences under matched content, which is crucial for distinguishing high-fidelity forgeries.
Meanwhile, despite realistic textures, synthetic videos may still violate physical or semantic logic. CoCoDetect therefore applies a confidence-gated strategy: uncertain cases are routed to an MLLM for reasoning about motion plausibility and scene coherence.
This framework combines contrastive learning with MLLM semantic verification, enhancing detection robustness across both texture-level and semantic-level forgery cues.

We validate our approach through extensive experiments on CoCoVideo and several open-source AIGC benchmarks, demonstrating state-of-the-art performance and strong robustness in cross-dataset generalization and ablation studies.

Our main contributions, as illustrated in Figure~\ref{fig:teaser}, are summarised as follows:
\begin{itemize}
\item We establish a high-quality video forgery detection dataset \textbf{CoCoVideo-26K} with contrastive real--fake video pairs sharing the same first frame and text description, covering multiple commercial video-generation models at a scale of 26K clips, and provide the research community with a new benchmark.

\item We introduce a novel detection framework \textbf{CoCoDetect} that combines contrastive learning and a confidence-gated MLLM-assisted mechanism, enabling systematic judgment of videos from both texture features and semantic features.

\item We conduct comprehensive experiments across our dataset and other public benchmarks, demonstrating the robustness and superiority of our approach.
\end{itemize}

\section{Related work}
\label{sec:relatedWork}

%-------------------------------------------------------------------------
\subsection{Forged Video Datasets}

Traditional deepfake detection datasets focused on facial manipulations using face-swap~\cite{deepfake2019, faceswap2018} and GAN-based methods~\cite{thies2016face2face, thies2019deferred}.
Representative datasets~\cite{rossler2019faceforensics++, li2020celeb, dolhansky2020deepfake, zi2020wilddeepfake} expanded forgery diversity through refined pipelines and post-processing.
While facial Deepfake datasets are abundant, diffusion-based T2V and I2V methods have advanced general video generation.
The development of such models was marked by the breakthrough of Sora~\cite{brooks2024video}, followed by numerous open-source~\cite{hong2022cogvideo, yuan2024mora, blattmann2023stable, khachatryan2023text2video, hacohen2024ltx, kong2024hunyuanvideo} and commercial models~\cite{jimengWeb, hailuoWeb, lumaWeb, pixverseWeb, runwayWeb, seedanceWeb, veoWeb, klingWeb, pikaWeb, viduWeb, vivagoWeb, midjourneyWeb, soraWeb}. 

Previous AIGC video forgery detection datasets~\cite{bai2024ai, ma2024decof, chen2024demamba, ni2025genvidbench, wen2025busterx} primarily rely on open-source models for content generation.
For instance, the latest benchmark BusterX~\cite{wen2025busterx} contains over 200K samples, yet only about 1,000 videos are sourced from commercial models.
Limited by the quality of open-source models, the generated videos differ significantly from real ones, making them easily distinguishable even by human eyes.

%-------------------------------------------------------------------------
\subsection{Traditional Detection Methods}

Early deepfake detection methods primarily employ CNN architectures to capture localized manipulation artifacts.
Representative works~\cite{rossler2019faceforensics++,zhao2021multi,nguyen2024laa, tan2024rethinking} design networks to detect facial texture inconsistencies and forgery traces.
For motion consistency, spatio-temporal CNN approaches~\cite{zheng2021exploring, guo2025face,nguyen2025vulnerability,zhang2024learning} reveal temporal anomalies in facial movements.
Frequency-based methods~\cite{kashiani2025freqdebias,kim2025beyond,tan2024frequency,wang2023dynamic} further enhance cross-model generalization by fusing frequency-domain features and correcting biases.
Concurrently, classic architectures such as EfficientNet~\cite{tan2019efficientnet}, ResNet~\cite{he2016deep}, and Xception~\cite{chollet2017xception} have become standard backbones for detection frameworks.
With diffusion-based generative models emerging, recent work targets AIGC-specific artifacts~\cite{song2024on}, temporal anomalies in large-scale videos~\cite{chen2024demamba}, and unified multimodal detection~\cite{kundu2025towards}.
However, these learning-based methods remain constrained by training on known generators and struggle to generalize to unseen generation techniques.

%-------------------------------------------------------------------------
\subsection{MLLM-Based Detection Methods}

LLMs such as GPT-4~\cite{achiam2023gpt}, LLaMA~\cite{touvron2023llama}, Qwen~\cite{bai2023qwen}, and Gemini~\cite{team2023gemini} have achieved remarkable success in natural language reasoning. Extending this capability to the visual domain, MLLMs integrate visual and textual modalities, with prominent examples including GPT-4o~\cite{hurst2024gpt}, Qwen2.5-VL~\cite{qwen25vl}, DeepSeek-VL~\cite{lu2024deepseekvl}, LLaVA~\cite{liu2023visual}, InternVL~\cite{chen2024internvl, chen2024far}, and Yi-VL~\cite{young2024yi}.
These MLLMs provide new opportunities for visual content authentication by compensating for traditional models' limitations in semantic understanding.

Consequently, the advanced reasoning capabilities of MLLMs are increasingly leveraged for forgery detection tasks.
Recent works~\cite{xu2024fakeshield,huang2025sida,fan2024fake,wen2025busterx} leverage MLLMs for detection, localization, and explanation, demonstrating enhanced generalization and interpretability.
However, these approaches primarily rely on tuning or prompting large models with insufficient integration of fine-grained network structures, restricting their ability to detect subtle low-level artifacts.

\section{Benchmark}
\label{sec:benchmark}

%-------------------------------------------------------------------------
\subsection{Data Sources and Composition}

The efficacy of AIGC video detection models is fundamentally constrained by the quality of their training data.
However, existing AIGC video datasets are mainly generated by open-source models, whose outputs remain limited in texture fidelity and semantic coherence compared to commercial counterparts.
Models trained on such data often fail to generalize to real-world scenarios, where high-quality commercial AIGC videos exhibit much stronger realism and diversity.
To bridge this gap and provide more reliable supervision for distinguishing real and synthetic content, we construct \textbf{CoCoVideo-26K}, a high-quality benchmark dataset built using multiple state-of-the-art commercial video generation models and organized into semantic-aligned real--fake video pairs.
Next, we introduce the composition of CoCoVideo.

\noindent\textbf{Real Videos (Original Videos):} 

To construct the real video subset of our dataset, we sourced high-quality real-world footage from the OpenVid-1M~\cite{nan2024openvid} dataset.
We meticulously filtered and selected approximately 13,000 videos, prioritizing high-fidelity, real-world scene data.
This stringent selection process is crucial for enabling the detection model to accurately perceive realistic scenes and effectively capture subtle genuine texture details.
These selected videos primarily feature a consistent duration of around 5 seconds, matching the default generation duration of most contemporary generative models, thereby ensuring robust temporal comparability between the real and synthetic video samples.
Furthermore, we leveraged the dataset's pre-existing textual descriptions associated with each video.
These descriptions were adopted as the direct prompts for our generation process, guaranteeing that every real--synthetic video pair maintains strong semantic consistency.

\begin{table}[t]
  \centering
  \small
  \vspace{-0.2em}
  \caption{Generation model version and their release date.}
  \vspace{-0.6em}
  \label{tab:dataset_composition}
  
  \begin{tabular*}{1.02\columnwidth}{@{\hspace{2mm}\extracolsep{\fill}}ccc@{\hspace{2mm}}}
    \toprule
    Platform & Model Version & Release Date \\
    \midrule
    Jimeng     & Jimeng 3.0~\cite{jimengWeb}                 & 2025-04 \\
    Hailuo     & MiniMax Hailuo 02~\cite{hailuoWeb}          & 2025-08 \\
    Luma       & Luma Ray Flash 2~\cite{lumaWeb}             & 2025-03 \\
    Pixverse   & Pixverse v5~\cite{pixverseWeb}              & 2025-08 \\
    Runway     & Runway Gen4 Turbo~\cite{runwayWeb}          & 2025-04 \\
    Seedance   & Seedance-1-0-lite-i2v~\cite{seedanceWeb}    & 2025-05 \\
    Veo        & Veo3 Fast~\cite{veoWeb}                     & 2025-08 \\
    Kling      & Kling 2.5 Turbo~\cite{klingWeb}             & 2025-09 \\
    Pika       & Pika 2.2~\cite{pikaWeb}                     & 2025-02 \\
    Vidu       & Vidu Q2~\cite{viduWeb}                      & 2025-09 \\
    Vivago     & Vivago 2.0~\cite{vivagoWeb}                 & 2025-06 \\
    Midjourney & Midjourney v7~\cite{midjourneyWeb}          & 2025-04 \\
    Sora       & Sora v1~\cite{soraWeb}                      & 2024-12 \\
    \bottomrule
  \end{tabular*}
  \vspace{-1.3em}
  
\end{table}

\noindent\textbf{Fake Videos (Generated Videos):} 
 
We synthesize the generated videos subset using a diverse selection of prominent commercial video generation models.
To ensure that our selection encompasses state-of-the-art generative models, we referenced platforms such as \textit{ArtificialAnalysis.ai}~\cite{aiModelWeb}.
These resources provide real-time metrics, including \textit{Model ELO rating}, \textit{95\% confidence intervals}, and \textit{appearance statistics}, enabling us to analyze various contemporary models from a multi-faceted perspective.
Based on these established rankings and our empirical observations, we selected 13 distinct commercial models.

We ensure balanced representation across different generation methodologies by having each model contribute exactly \textbf{1,000} videos to the final dataset composition.
These selected models exhibit significant advancements over prevailing open-source methods, particularly regarding visual fidelity, inter-frame motion consistency, and precise semantic control.
The specific version details and availability dates for each model are fully documented in Table~\ref{tab:dataset_composition}.

%-------------------------------------------------------------------------

\begin{table*}[t]
  \centering
  \caption{Comparison of CoCoVideo with existing AIGC video datasets.
}
  \label{tab:comparison}
  \begin{tabular*}{0.95\textwidth}{@{\hspace{2mm}\extracolsep{\fill}}ccccccc@{\hspace{2mm}}}
    \toprule
    Dataset & Latest Model & Scale & \makecell{Model Source} & FPS & Aligned Pairs & Modality \\
    \midrule
    GVD~\cite{bai2024ai}           
    & 2024-01       & 11K    & Mixed     & -     & \xmark & Video  \\
    GVF~\cite{ma2024decof}           
    & 2024-03       & 2.8K   & Mixed     & -     & \xmark & Video  \\
    GenVideo~\cite{chen2024demamba}      
    & 2024-04       & 2,271K & Mixed     & 8-24  & \xmark & Video  \\
    GenVidBench~\cite{ni2025genvidbench}   
    & 2024-04       & 143K   & Mixed     & 3-30  & \xmark & Video  \\
    GenBuster~\cite{wen2025busterx}     
    & 2025-01       & 200K   & Mixed     & 24    & \xmark & Video  \\
    \midrule
    \textbf{CoCoVideo(Ours)} 
    & 2025-09  & 26K    & Commercial   & 24-30 & \cmark & Video \& Text \\
    \bottomrule
  \end{tabular*}
  \vspace{-0.5em}
\end{table*}

\subsection{Dataset Construction Pipeline}
\label{datasetConstructionPipeline}

\begin{figure}[t]
  \centering
   \includegraphics[width=1.0\linewidth]{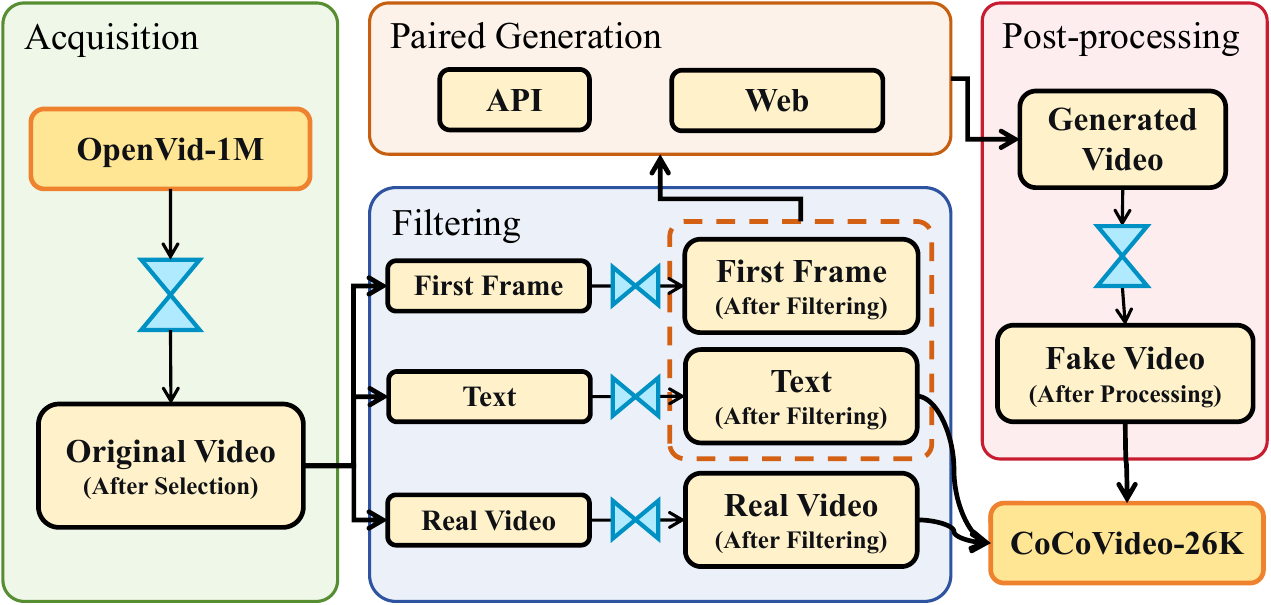}
   \caption{CoCoVideo construction pipeline. (1) \textit{Acquisition} from OpenVid-1M, selecting diverse original videos; (2) \textit{Filtering} by first-frame quality, text prompt suitability, and real video standards; (3) \textit{Paired Generation} via API or web interfaces to produce semantically aligned fake videos; (4) \textit{Post-processing} to obtain strictly aligned real--fake pairs with comparable properties.}
   \label{fig:construction}
   \vspace{-1.0em}
\end{figure}

As illustrated in Figure~\ref{fig:construction}, we constructed our CoCoVideo dataset with a four-stage pipeline: acquisition, filtering, paired generation, and post-processing with archiving.
This pipeline ensures that each real--fake video pair is strictly aligned in semantic description and first-frame condition, while remaining comparable in duration and resolution.
To enhance the comprehensiveness of our dataset, we include not only basic talking-heads and human-interaction content but also additional categories such as food preparation, plant growth, cultural architecture, and natural landscapes.

The specific implementation details of these stages, as well as the detailed composition of the dataset are described in the supplementary material.

%-------------------------------------------------------------------------

\subsection{Comparison with Other Benchmarks}

As shown in Table~\ref{tab:comparison}, we systematically compare our constructed dataset with several mainstream publicly available AIGC video datasets, focusing on dimensions such as dataset scale, the latest model available time, main generative model source, etc.

Our dataset's distinct advantage over existing benchmarks lies primarily in its data sourcing and structured design. While existing datasets often rely heavily on open-source models or mix them with older commercial versions, CoCoVideo strategically incorporates synthetic data generated by a diverse collection of novel commercial generative models.
This critical distinction ensures \textbf{high-quality} samples with enhanced real-world representativeness.
Moreover, to ensure overall consistency, our dataset enforces a unified video duration and maintains approximately consistent frame rates across all samples.
Importantly, to further broaden its applicability and enable diverse research paradigms in AIGC video detection, CoCoVideo-26K provides structurally contrastive real--fake video pairs accompanied by their corresponding text prompts.
This multimodal design enriches the dataset’s semantic diversity and enables future researchers to make consistent and fair comparisons using identical generative prompts.

\section{Method}
\label{sec:method}

%-------------------------------------------------------------------------

\begin{figure*}[t]
  \centering
   \includegraphics[width=0.9\linewidth]{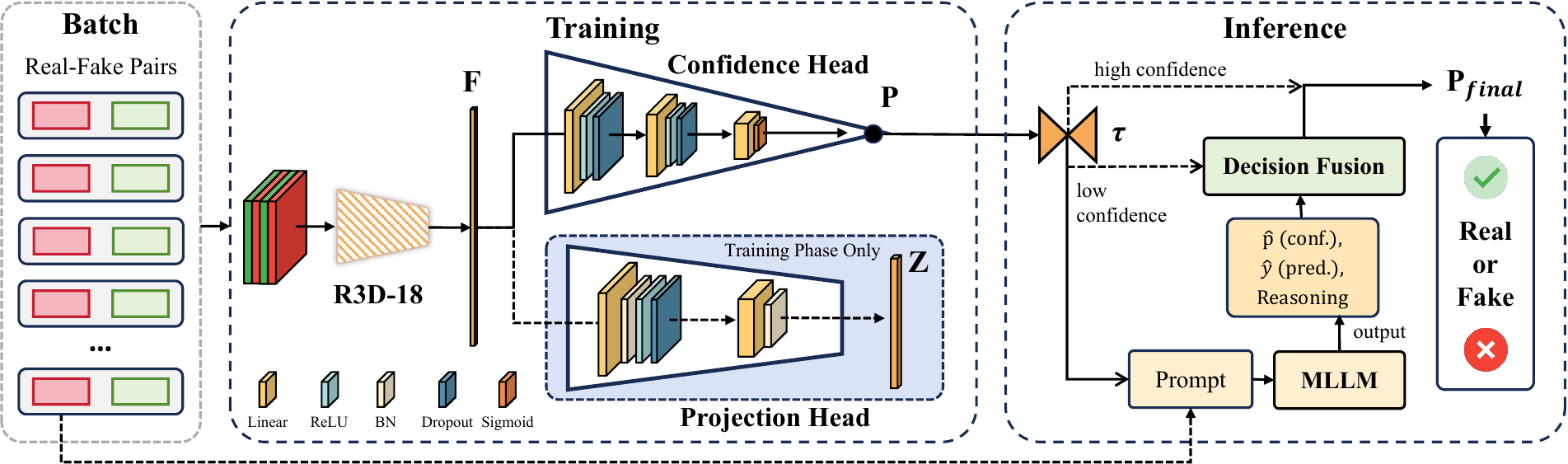}
   \caption{CoCoDetect pipeline. Training stage processes real--fake video pairs: R3D-18 extracts features $\mathbf{F}$, fed into dual heads producing confidence $\mathbf{P}$ (for authenticity classification) and contrastive embeddings $\mathbf{Z}$ (for paired contrastive loss). Inference stage uses threshold $\tau$ to route uncertain samples to MLLM, which outputs structured JSON \{$\hat{y}$, $\hat{p}$, reasoning\}. The MLLM outputs are then fused with $\mathbf{P}$ via decision fusion to yield $P_{\text{final}}$ for final classification.}
   \label{fig:pipeline}
   \vspace{-1.0em}
\end{figure*}

In this section, we present \textbf{CoCoDetect}, a novel architecture specifically tailored for the authenticity discrimination of AIGC videos. Our framework is built on an R3D-18 spatiotemporal backbone~\cite{he2016deep, hara2017learning, tran2018closer, hara2018can} with a dual-head design and a confidence-gated strategy. Crucially, our CoCoDetect leverages the semantically well-aligned real--fake video pairs from our CoCoVideo dataset to facilitate stable and highly generalizable model learning.
In the following subsections we will introduce each part of our proposed framework.

%-------------------------------------------------------------------------

\subsection{Dual-Head Contrastive Training}

As Figure~\ref{fig:pipeline} illustrates, we build our training pipeline upon paired video data, extract features through a spatiotemporal convolutional backbone, and employ a dual-head architecture to jointly optimize classification capability and representation discriminability. 

\subsubsection{Data Organization}

The training phase adopts a paired batch construction strategy.
Each batch contains $B=2N_{\text{pairs}}$ video samples, composed of $N_{\text{pairs}}$ real--fake pairs.
For the $i$-th sample $v_i$ ($i=1,2,\ldots,B$), three key pieces of information are maintained: \textbf{binary label} $\bm{y_i} \in \{0,1\}$; \textbf{pair index} $\bm{\pi_i} \in \{1,2,\ldots,N_{\text{pairs}}\}$, which identifies the video pair to which the sample belongs; and \textbf{video tensor} $\bm{v_i} \in \mathbb{R}^{C \times T \times H \times W}$, where $C=3$ denotes RGB channels, $T$ represents the temporal dimension, and $H \times W$ is the spatial resolution.

For any pair of real--fake videos satisfying $\pi_i = \pi_j$ and $y_i \neq y_j$, both videos share the same first frame and text description, achieving strong semantic alignment.

\subsubsection{Network Architecture}

We adopt R3D-18 as the spatiotemporal feature extraction backbone.
For an input batch $\mathbf{X} \in \mathbb{R}^{B \times C \times T \times H \times W}$, the backbone outputs a feature matrix:
\begin{equation}
\mathbf{F} = f_{\text{backbone}}(\mathbf{X}), \mathbf{F} \in \mathbb{R}^{B \times d_{\text{feat}}}
%\mathbf{F} = f_{\text{backbone}}(\mathbf{X}) \in \mathbb{R}^{B \times d_{\text{feat}}}
\end{equation}
where $d_{\text{feat}}$ denotes the feature dimension (512 by default for R3D-18).

Next, the feature matrix $\mathbf{F}$ is fed into two parallel head networks:

\textbf{1) Confidence Head} employs a three-layer $\text{MLP}_{\text{conf}}$ followed by a Sigmoid activation function to produce a confidence score in the range $[0,1]$:
\begin{equation}
\mathbf{P} = \sigma(\text{MLP}_{\text{conf}}(\mathbf{F})), \mathbf{P} \in [0,1]^B
\end{equation}
where $\mathbf{P}$ is the batch confidence vector, with $p_i$ representing the authenticity confidence of the $i$-th sample: higher $p_i$ indicates stronger confidence in being real, while lower $p_i$ suggests the sample is generated.

\textbf{2) Projection Head} uses a two-layer $\text{MLP}_{\text{proj}}$ with batch normalization between layers to map the backbone-extracted features $\mathbf{F}$ into a 128-dimensional contrastive embedding space:
\begin{equation}
\mathbf{Z} = \frac{\text{MLP}_{\text{proj}}(\mathbf{F})}{\|\text{MLP}_{\text{proj}}(\mathbf{F})\|_2}, \quad \mathbf{Z} \in \mathbb{R}^{B \times 128}
\end{equation}
where $\mathbf{Z}$ is the $\ell_2$-normalized embedding matrix (each row $\mathbf{z}_i$ corresponds to the normalized output of $\text{MLP}_{\text{proj}}$) and is used to compute the paired contrastive loss.

\subsubsection{Loss Function Design}

The total training loss is given by $\mathcal{L}_{\text{total}} = \alpha \cdot \mathcal{L}_{\text{conf}} + (1 - \alpha) \cdot \mathcal{L}_{\text{pair}}$, where $\alpha$ is the weight balancing the contributions of classification accuracy and contrastive discriminability.

\paragraph{Confidence Loss:}
The confidence loss $\mathcal{L}_{\text{conf}}$ employs binary cross-entropy (BCE)  loss to optimize authenticity prediction:
\begin{equation}
\mathcal{L}_{\text{conf}} = -\frac{1}{B} \sum_{i=1}^{B} \left[ y_i \log(p_i) + (1-y_i) \log(1-p_i) \right]
\end{equation}

\paragraph{Paired Contrastive Loss:}
Traditional supervised contrastive learning enforces intra-class compactness by pulling together all samples of the same label while separating different classes.
However, this approach struggles in AIGC detection: real videos span diverse scenes and content, while generated videos vary across different generation methods.
Such high intra-class semantic variance makes global class-level clustering counterproductive.

Instead, CoCoDetect focuses on paired discriminability: we only enforce separation between semantically aligned real--fake pairs that share the same first frame and text prompt, without imposing constraints on non-paired samples.
Specifically, for each real--fake pair $(i, j)$ where $\pi_i = \pi_j$ and $y_i \neq y_j$, we compute the cosine similarity as $s_{ij} = \mathbf{z}_i \cdot \mathbf{z}_j$.
The loss applies a hinge penalty when similarity exceeds the target threshold:
\begin{equation}
\mathcal{L}_{\text{pair}} = \frac{1}{N_{\text{pairs}}} \sum_{\substack{i,j=1}}^{B} \max(0, s_{ij} - (1 - m))
\end{equation}
where the summation is over all $N_{\text{pairs}}$ real--fake pairs $(i,j)$ satisfying $\pi_i = \pi_j$ and $y_i \neq y_j$, and $m \in [0.5,1.5]$ is a margin hyperparameter controlling the target separation degree.
When $m$ increases from $0.5$ to $1.5$, the cosine-threshold decreases from $0.5$ to $–0.5$, which corresponds to the embedding separation degree strengthening from small angular distance ($60^\circ$) to a large one ($120^\circ$).
It means a stronger constraint on the separation between real--fake pair embeddings.

%-------------------------------------------------------------------------

\subsection{Confidence-Gated MLLM-Assisted Inference}

As shown in Figure~\ref{fig:pipeline}, during the inference phase, the base network processes the input video to output a confidence score $p \in [0,1]$.
Based on the threshold $\tau$, the confidence-gated mechanism directly outputs prediction results for high-confidence samples, while uncertain samples are routed to the MLLM for further reasoning.

\subsubsection{Confidence-Gated Mechanism}

The confidence-gated mechanism is motivated by a key empirical observation: model prediction accuracy is strongly correlated with confidence scores.
Although this phenomenon is common across machine learning tasks, prior forgery detection methods rarely exploit this correlation due to the lack of effective alternative reasoning pathways for low-confidence cases.
Fortunately, the development of MLLMs provides a new perspective for forged video detection tasks: unlike CNN-based models, which focus on capturing spatiotemporal texture features, we can utilize the semantic understanding capability of MLLMs to reason based on the physical plausibility and scene coherence of video content, thereby achieving comprehensive discrimination of forged videos.

Through systematic ablation experiments (see Section~\ref{sec:ablationStudies}), we evaluated the accuracy of the confidence head and MLLM across different confidence ranges, and identified the final confidence threshold value $\tau$.

\subsubsection{MLLM-Assisted Reasoning}
For low-confidence samples routed to MLLM ($1-\tau < p < \tau$), we design a structured reasoning pipeline with three key stages, which are detailed as follows:

\noindent\textbf{Prompt Construction.}
The MLLM input consists of three complementary components.
\textbf{1) Video content:} the MLLM automatically extracts a sequence of key frames covering the temporal span based on a preset sampling strategy.
\textbf{2) Confidence score $p$:} derived from the base model's confidence head output, informing the MLLM of the sample's uncertainty level.
\textbf{3) Textual reasoning instructions:} explicitly guiding the MLLM to analyze physical plausibility, temporal consistency, and scene coherence in the video, while defining the task, integrating video and confidence information, and standardizing the MLLM's structured output format.

\noindent\textbf{Structured Output Design.}
To ensure effective integration of the MLLM's analysis with the base model's predictions, we design a three-part structured JSON output format and require the model to strictly output results in JSON format.
\textbf{1) Prediction $\hat{y}\in\{0,1\}$:} the MLLM's binary judgment (fake or real). 
\textbf{2) Confidence $\hat{p}\in[0,1]$:} the MLLM's semantic certainty score reflecting its confidence level grounded in commonsense reasoning.
\textbf{3) Reasoning:} a natural language explanation describing the basis for the judgment result and certainty level.
Those outputs that fail to meet the JSON format requirements are recorded and serve as one of the criteria for MLLM model selection.

\noindent\textbf{Decision Fusion Strategy:}
Since the MLLM's confidence $\hat{p}$ has different semantics from the base model's confidence $p$, we first map $\hat{p}$ to the same semantic space as $p$:
\begin{equation}
\tilde{p} = 0.5 \left( \hat{y}(1 + \hat{p}) + (1 - \hat{y})(1 - \hat{p}) \right)
\end{equation}
where $\tilde{p}$ represents the MLLM's confidence, aligning with the semantics of the base model's $p$.
Subsequently, the final fused confidence $p_{\text{final}}$ is computed through adaptive weighting:
\begin{equation}
p_{\text{final}} = u \cdot \tilde{p} + (1 - u) \cdot p
\end{equation}
where the adaptive weight $u$ is determined by:
\begin{equation}
u = \sqrt{2|\tilde{p} - 0.5|}
\end{equation}
% When $\tilde{p}$ approaches 0.5, $u$ decreases, reducing the MLLM's contribution; when $\tilde{p}$ is far from 0.5, $u$ increases rapidly due to the square root amplification, prioritizing the MLLM's confident semantic judgment.
Initially, we explored learning the optimal value $u$ through training, but this led to over-reliance on MLLM outputs and degraded generalization.
Instead, this simple and fixed formula appropriately weights the MLLM's confidence while maintaining robustness across diverse benchmarks.
\section{Experiments}
\label{sec:experiments}

%-------------------------------------------------------------------------

\subsection{Implementation Details}
\label{experimentalSetup}

We employ LLaVA-NeXT-Video-7B~\cite{liu2024llavanext} as our reasoning MLLM.
During the training phase, we set the loss weight to $\alpha=0.65$, the margin parameter in the paired contrastive loss to $m = 1.0$, and the threshold of the confidence-gated inference mechanism to $\tau = 0.9$. The experiments are implemented on a single NVIDIA A6000 GPU with batch size of 8 and temporal dimension $T=16$. We use the AdamW optimizer with initial learning rate of $10^{-4}$ and weight decay of $10^{-4}$. The training process lasted 30 epochs, consuming approximately 10 hours.
Detailed experimental setups can be found in the supplementary material.

%-------------------------------------------------------------------------

\begin{table*}[t]
  \centering
  \caption{Comparison results on CoCoVideo test set \textbf{(\%)}. \textbf{Bold} and \underline{underlined} scores indicate the 1st- and 2nd-ranked performances.}
   \renewcommand{\arraystretch}{1.1}
  \label{tab:cocovideo_results}
  \resizebox{1.0\textwidth}{!}{
    \newcolumntype{W}{>{\centering\arraybackslash}p{1.1cm}}
      % \begin{tabular}{c|c|ccccccccccccc|c}
      \begin{tabular}{c|c|WWWWWWWWWWWWW|c}
        \toprule
        \multicolumn{1}{c|}{\multirow{2}{*}{Methods}} & 
        \multicolumn{1}{c|}{\multirow{2}{*}{Metric (\%)}} & 
        \multicolumn{13}{c|}{Models} &
        \multicolumn{1}{c}{\multirow{2}{*}{Avg.}} \\ 
        % \cline{3-15} 
        \multicolumn{1}{c|}{} & 
        \multicolumn{1}{c|}{} & 
        Jimeng & Hailuo & Luma & Pixverse & Runway & Seedance & Veo3 & Kling & Pika & Vidu & Vivago & Midjourney & Sora \\ 
        
        \midrule
        
        \multicolumn{1}{c|}{\multirow{4}{*}{3D ResNet~\cite{hara2018can}}} & \multicolumn{1}{c|}{Acc} & 74.00 & 80.33 & 78.00 & \underline{87.33} & 79.33 & 81.67 & 86.33 & 66.00 & 76.67 & 70.67 & 70.00 & 88.33 & \multicolumn{1}{c|}{84.33} & 78.69 \\ 
        \multicolumn{1}{c|}{} & \multicolumn{1}{c|}{F1} & 76.79 & 80.40 & 75.56 & \underline{87.16} & 78.91 & 81.10 & 86.29 & 65.77 & 77.27 & 70.86 & 69.80 & 87.80 & \multicolumn{1}{c|}{84.18} & 78.57 \\ 
        \multicolumn{1}{c|}{} & \multicolumn{1}{c|}{Recall} & 86.00 & \underline{80.67} & 68.00 & 86.00 & 77.33 & 78.67 & 86.00 & 65.33 & 79.33 & 71.33 & 69.33 & 84.00 & \multicolumn{1}{c|}{83.33} & 78.10 \\ 
        \multicolumn{1}{c|}{} & \multicolumn{1}{c|}{AUC} & 85.19 & 87.60 & 87.50 & \underline{94.46} & 87.81 & 90.17 & 94.15 & 69.85 & 83.71 & 76.72 & 74.85 & 94.69 & \multicolumn{1}{c|}{92.11} & 86.51 \\ 
        
        \midrule
        
        \multicolumn{1}{c|}{\multirow{4}{*}{3D ResNeXt~\cite{hara2018can}}} & \multicolumn{1}{c|}{Acc} & 75.00 & 83.00 & 79.00 & 82.33 & 73.67 & 83.00 & 82.00 & \underline{77.67} & \underline{83.67} & \underline{84.00} & \underline{79.33} & 79.33 & \multicolumn{1}{c|}{84.33} & 80.49 \\ 
        \multicolumn{1}{c|}{} & \multicolumn{1}{c|}{F1} & 79.56 & 83.39 & 78.35 & 82.27 & 73.58 & 83.17 & 81.88 & \underline{77.59} & \underline{83.83} & \underline{83.45} & 78.47 & 78.91 & \multicolumn{1}{c|}{84.07} & 80.65 \\ 
        \multicolumn{1}{c|}{} & \multicolumn{1}{c|}{Recall} & \textbf{97.33} & \textbf{85.33} & \underline{76.00} & 82.00 & 73.33 & 84.00 & 81.33 & 77.33 & 84.67 & 80.67 & 75.33 & 77.33 & \multicolumn{1}{c|}{82.67} & 81.33 \\ 
        \multicolumn{1}{c|}{} & \multicolumn{1}{c|}{AUC} & \underline{89.14} & 90.44 & 85.71 & 91.44 & 82.88 & 88.45 & 90.35 & 84.47 & 91.01 & 91.06 & 88.12 & 87.01 & \multicolumn{1}{c|}{92.03} & 88.16 \\ 
        
        \midrule
        
        \multicolumn{1}{c|}{\multirow{4}{*}{VideoMAE~\cite{tong2022videomae}}} & \multicolumn{1}{c|}{Acc} & 73.33 & 77.33 & 82.67 & 84.33 & 74.67 & 86.67 & 90.33 & 57.33 & 74.00 & 81.00 & 79.00 & 86.00 & \multicolumn{1}{c|}{\underline{88.67}} & 79.64 \\ 
        \multicolumn{1}{c|}{} & \multicolumn{1}{c|}{F1} & 76.05 & 73.64 & 81.02 & 82.78 & 71.43 & 85.71 & 90.17 & 42.86 & 71.74 & 79.72 & 77.26 & 84.78 & \multicolumn{1}{c|}{\underline{88.36}} & 77.98 \\ 
        \multicolumn{1}{c|}{} & \multicolumn{1}{c|}{Recall} & 84.67 & 63.33 & 74.00 & 75.33 & 63.33 & 80.00 & 88.67 & 32.00 & 66.00 & 74.67 & 71.33 & 78.00 & \multicolumn{1}{c|}{\underline{86.00}} & 72.10 \\ 
        \multicolumn{1}{c|}{} & \multicolumn{1}{c|}{AUC} & 83.98 & \underline{90.94} & 91.53 & 93.91 & 87.19 & 95.26 & 96.75 & 62.86 & 86.83 & 89.37 & 84.48 & 94.63 & \multicolumn{1}{c|}{\underline{95.07}} & 89.26 \\ 
        
        \midrule
        
        \multicolumn{1}{c|}{\multirow{4}{*}{TALL~\cite{xu2023tall}}} & \multicolumn{1}{c|}{Acc} & 66.67 & 68.67 & 74.00 & 76.67 & 74.00 & 84.33 & 87.67 & 50.33 & 68.00 & 62.00 & 62.00 & 74.00 & \multicolumn{1}{c|}{74.33} & 70.97 \\ 
        \multicolumn{1}{c|}{} & \multicolumn{1}{c|}{F1} & 67.74 & 61.48 & 69.77 & 72.66 & 70.68 & 83.62 & 87.71 & 36.05 & 64.71 & 58.70 & 56.82 & 68.80 & \multicolumn{1}{c|}{69.32} & 67.36 \\ 
        \multicolumn{1}{c|}{} & \multicolumn{1}{c|}{Recall} & 70.00 & 50.00 & 60.00 & 62.00 & 62.67 & 80.00 & 88.00 & 28.00 & 58.67 & 54.00 & 50.00 & 57.33 & \multicolumn{1}{c|}{58.00} & 59.90 \\ 
        \multicolumn{1}{c|}{} & \multicolumn{1}{c|}{AUC} & 74.83 & 82.08 & 84.51 & 86.16 & 80.58 & 92.38 & 94.49 & 49.75 & 76.84 & 69.16 & 66.82 & 86.57 & \multicolumn{1}{c|}{86.75} & 80.51 \\ 
        
        \midrule
        
        \multicolumn{1}{c|}{\multirow{4}{*}{D3~\cite{zheng2025d3}}} & \multicolumn{1}{c|}{Acc} & 49.00 & 47.67 & 49.33 & 48.33 & 50.00 & 50.67 & 50.00 & 53.33 & 48.67 & 49.67 & 48.67 & 44.33 & \multicolumn{1}{c|}{46.67} & 48.95 \\ 
        \multicolumn{1}{c|}{} & \multicolumn{1}{c|}{F1} & 9.47 & 14.21 & 16.48 & 19.69 & 14.77 & 22.11 & 13.79 & 16.67 & 2.53 & 1.31 & 3.75 & 4.57 & \multicolumn{1}{c|}{4.76} & 11.47 \\ 
        \multicolumn{1}{c|}{} & \multicolumn{1}{c|}{Recall} & 5.33 & 8.67 & 10.00 & 12.67 & 8.67 & 14.00 & 8.00 & 9.33 & 1.33 & 0.67 & 2.00 & 2.67 & \multicolumn{1}{c|}{2.67} & 6.62 \\ 
        \multicolumn{1}{c|}{} & \multicolumn{1}{c|}{AUC} & 61.09 & 48.74 & 53.03 & 50.37 & 53.53 & 60.41 & 54.48 & 62.97 & 37.15 & 45.48 & 17.19 & 36.50 & \multicolumn{1}{c|}{40.00} & 48.40 \\ 
        
        \midrule

        \multicolumn{1}{c|}{\multirow{4}{*}{DeMamba~\cite{chen2024demamba}}} & \multicolumn{1}{c|}{Acc} & \textbf{83.00} & \underline{84.33} & \underline{80.67} & 87.00 & \underline{84.00} & \underline{89.33} & \textbf{94.00} & 71.33 & 80.67 & 79.00 & 74.00 & \underline{89.67} & \multicolumn{1}{c|}{76.67} & \underline{82.59} \\ 
        \multicolumn{1}{c|}{} & \multicolumn{1}{c|}{F1} & \textbf{83.71} & \underline{83.62} & \underline{78.52} & 86.69 & \underline{84.11} & \underline{89.54} & \textbf{94.16} & 77.25 & 83.71 & 82.64 & \underline{79.03} & \underline{89.56} & \multicolumn{1}{c|}{73.88} & \underline{83.49} \\ 
        \multicolumn{1}{c|}{} & \multicolumn{1}{c|}{Recall} & 87.33 & 80.00 & 70.67 & \underline{84.67} & \textbf{84.67} & \textbf{91.33} & \textbf{96.67} & \textbf{97.33} & \textbf{99.33} & \textbf{100.00} & \textbf{98.00} & \underline{88.67} & \multicolumn{1}{c|}{66.00} & \underline{88.05} \\ 
        \multicolumn{1}{c|}{} & \multicolumn{1}{c|}{AUC} & \textbf{90.22} & 90.80 & \underline{88.38} & 92.46 & \underline{90.65} & \textbf{95.92} & \textbf{98.06} & \underline{92.64} & \underline{97.96} & \underline{98.93} & \underline{94.29} & \underline{96.35} & \multicolumn{1}{c|}{84.75} & \underline{91.03} \\ 
        
        \midrule
        
        \multicolumn{1}{c|}{\multirow{4}{*}{\makecell{CoCoDetect\\(Ours)}}} & \multicolumn{1}{c|}{Acc} & \underline{78.00} & \textbf{86.00} & \textbf{89.67} & \textbf{93.33} & \textbf{85.00} & \textbf{91.00} & \underline{93.33} & \textbf{91.67} & \textbf{94.67} & \textbf{96.67} & \textbf{94.33} & \textbf{95.00} & \multicolumn{1}{c|}{\textbf{90.33}} & \textbf{90.69} \\ 
        \multicolumn{1}{c|}{} & \multicolumn{1}{c|}{F1} & \underline{80.00} & \textbf{84.89} & \textbf{89.12} & \textbf{93.10} & \textbf{84.21} & \textbf{90.53} & \underline{93.42} & \textbf{91.58} & \textbf{94.81} & \textbf{96.75} & \textbf{94.53} & \textbf{94.98} & \multicolumn{1}{c|}{\textbf{90.03}} & \textbf{90.62} \\ 
        \multicolumn{1}{c|}{} & \multicolumn{1}{c|}{Recall} & \underline{88.00} & 78.67 & \textbf{84.67} & \textbf{90.00} & \underline{80.00} & \underline{86.00} & \underline{94.67} & \underline{90.67} & \underline{97.33} & \underline{99.33} & \textbf{98.00} & \textbf{94.67} & \multicolumn{1}{c|}{\textbf{87.33}} & \textbf{89.95} \\ 
        \multicolumn{1}{c|}{} & \multicolumn{1}{c|}{AUC} & 88.54 & \textbf{93.35} & \textbf{94.10} & \textbf{98.16} & \textbf{91.27} & \underline{95.87} & \underline{97.94} & \textbf{97.02} & \textbf{98.56} & \textbf{99.88} & \textbf{97.78} & \textbf{98.77} & \multicolumn{1}{c|}{\textbf{95.30}} & \textbf{95.93} \\

        \bottomrule
      \end{tabular}
  }
  \label{tab:performance_comparison}
  \vspace{-0.5em}
\end{table*}

\subsection{Comparative Experiments}
\subsubsection{Performance on CoCoVideo Dataset}

As shown in Table~\ref{tab:cocovideo_results}, we evaluate the performance of CoCoDetect on our proposed CoCoVideo test set and benchmark it against a diverse set of open-source methods, including 3D ResNet~\cite{hara2018can}, 3D ResNeXt~\cite{hara2018can}, VideoMAE~\cite{tong2022videomae}, TALL~\cite{xu2023tall}, D3~\cite{zheng2025d3}, and DeMamba~\cite{chen2024demamba}.
For a comprehensive comparison, we report detailed statistics for each generative model within the dataset.
All competing methods (except D3, which is a training-free approach) are trained using the same training set and setup described in Section~\ref{experimentalSetup}.
The comparison encompasses four key metrics: Accuracy (ACC), F1-score (F1), Recall, and the Area under the ROC Curve (AUC).

Overall, our method achieves the best average performance across all four metrics, with DeMamba ranking second.
Several details merit further discussion.
From the \textbf{metric perspective}, our method achieves suboptimal Recall on 8 out of 13 models, yet maintains a higher F1-score, reflecting a conservative decision strategy that prioritizes precision.
This demonstrates that our confidence-gated fusion mechanism selectively integrates MLLM reasoning only for uncertain samples, effectively reducing false alarms while accepting some missed detections.
From the \textbf{model perspective}, our method does not achieve optimal results on Jimeng and Veo3.
Dataset analysis reveals that these generation models expand input content to match their predefined resolutions, introducing resolution inconsistencies between real--fake video pairs.
This reveals our method's reliance on semantically consistent real--fake pairs: performance degrades when semantic alignment is disrupted by factors such as resolution discrepancies.

\subsubsection{Cross-Dataset Generalization}

Most dataset-specific research trained and evaluated methods exclusively only on their own benchmarks.
This practice restricts rigorous assessment of methods’ generalization capabilities across diverse data distributions.
We select three strong-performing methods, 3D ResNeXt~\cite{hara2018can}, VideoMAE~\cite{tong2022videomae}, and DeMamba~\cite{chen2024demamba}, from our prior evaluation on CoCoVideo and compare them with our method on unseen datasets.
Using the same training data and setup ensures that the results reflect the varying generalization capabilities of different methods when confronted with unseen generation models.

As shown in Table~\ref{tab:cross_dataset}, we conduct evaluations across five public AIGC video benchmarks: GVD~\cite{bai2024ai} (test split, 11,068 clips), GVF~\cite{ma2024decof} (test split, 4,206 clips), GenVideo~\cite{chen2024demamba} (validation split, 18,286 clips), GenVidBench~\cite{ni2025genvidbench} (Pair1 subset, 71,501 clips) and GenBuster~\cite{wen2025busterx} (benchmark subset, 2,000 clips).
We report Accuracy and F1-score to assess cross-dataset generalization.
Notably, F1-score is omitted for GVD and GVF, as their test sets only contain fake videos, making this metric inapplicable.
Due to the varying video counts and class distributions across datasets, we do not report averaged scores.

Our method achieves high accuracy across all benchmarks, with substantial leads on the first four datasets.
Two phenomena merit analysis.
\textbf{First}, on GenVideo, despite achieving higher accuracy than others, F1-score drops to 33.19\% due to imbalanced performance: 84.8\% on real videos versus 23.55\% on fake videos. Investigation reveals that GenVideo's fake videos predominantly consist of short clips.
Our 16-frame requirement necessitates tail frame repetition, disrupting temporal coherence for motion analysis.
Additionally, these nearly static videos lack meaningful motion patterns, making it difficult for MLLMs to extract semantic cues and limiting our reasoning-based fusion effectiveness.
\textbf{Second}, our performance decreases on GenBuster.
Analysis shows that GenBuster employs earlier versions of the same models used in CoCoVideo, enabling methods to exploit generation-specific artifacts.
Conversely, our lower reliance on such patterns demonstrates stronger generalization across diverse benchmarks.

\begin{table}[t]
  \centering
  \caption{Comparison results on open-source benchmarks \textbf{(\%)}. \textit{GenV.} denotes GenVideo, \textit{GenV.B.} denotes GenVidBench, and \textit{GenB.} denotes GenBuster. \textbf{Bold} indicates best performance.}
     \vspace{-0.5em}
  \renewcommand{\arraystretch}{1.1}
  \label{tab:cross_dataset}
  \resizebox{1.0\linewidth}{!}{
    \begin{tabular}{c|c|ccccc}
      \toprule
      \multirow{2}{*}{Methods} &
      \multirow{2}{*}{Metric}  &
      \multicolumn{5}{c}{Benchmarks} \\
       & & GVD & GVF & GenV. & GenV.B. & GenB. \\
      \midrule
      
      \multirow{2}{*}{3DResNext~\cite{hara2018can}} 
      & Acc   & 45.20 & 33.83 & 35.20 & 51.90 & 35.80 \\
       & F1    & - & - & 37.76 & 42.39 & 37.73 \\
       
      \midrule
      
      \multirow{2}{*}{VideoMAE~\cite{tong2022videomae}} 
      & Acc   & 46.38 & 29.74 & 27.22 & 36.99 & \textbf{68.20} \\
      & F1    & - & - & 41.58 & 26.94 & \textbf{66.77} \\
      
      \midrule
      
      \multirow{2}{*}{DeMamba~\cite{chen2024demamba}} 
      & Acc   & 24.11 & 26.91 & 48.17 & 32.62 & 67.35 \\
      & F1    & - & - & \textbf{57.31} & 18.56 & 64.14 \\
      
      \midrule
      
      \multirow{2}{*}{\makecell{CoCoDetect\\(Ours)}} 
      & Acc   & \textbf{71.49} & \textbf{74.56} & \textbf{57.04} & \textbf{64.78} & 64.80 \\
      & F1    & - & - & 33.19 & \textbf{73.53} & 63.93 \\
      
      \bottomrule
    \end{tabular}
  }
  \vspace{-1.3em}
\end{table}

%-------------------------------------------------------------------------

\subsection{Ablation Study}
\label{sec:ablationStudies}

We perform a component ablation to verify the contribution of each module in CoCoDetect.
As shown in Table~\ref{tab:component_ablation}, four variants are evaluated in Accuracy, F1-score, and AUC:
\textbf{1) Backbone only:} 3D ResNet with confidence head.
\textbf{2) w/o Projection Head:} Full pipeline minus the paired contrastive projection branch.
\textbf{3) w/o MLLM:} Dual-head training but inference relies solely on the confidence score.
\textbf{4) CoCoDetect (Full):} Complete system with projection head and confidence-gated MLLM reasoning.

The full model attains the highest scores across all three metrics, showing that low-level contrastive features and high-level semantic reasoning work together to enhance overall detection robustness.

\begin{table}[ht]
    \centering
    \caption{Component ablation study \textbf{(\%)}. \textbf{Bold} indicates best performance.}
    \vspace{-0.5em}
    \label{tab:component_ablation}
    \begin{tabular*}{1.02\columnwidth}{@{\hspace{2mm}}@{\extracolsep{\fill}}lccc@{\hspace{2mm}}}
    \toprule
    \makecell{Model Variant} & \makecell{Acc} & \makecell{F1} & \makecell{AUC} \\
    \midrule
    Backbone only & 78.69 & 78.57 & 86.51\\
    w/o Projection Head & 81.05 & 83.17 & 91.39\\
    w/o MLLM & 88.92 & 88.79 & 95.46\\
    \midrule
    CoCoDetect (Full) & \textbf{90.69} & \textbf{90.62} & \textbf{95.93}\\
    \bottomrule
    \end{tabular*}
    \vspace{-1.0em}
\end{table}

\subsection{Case Study}

\begin{figure}[ht]
  \centering
  \vspace{-0.3em}
   \includegraphics[width=1.0\linewidth]{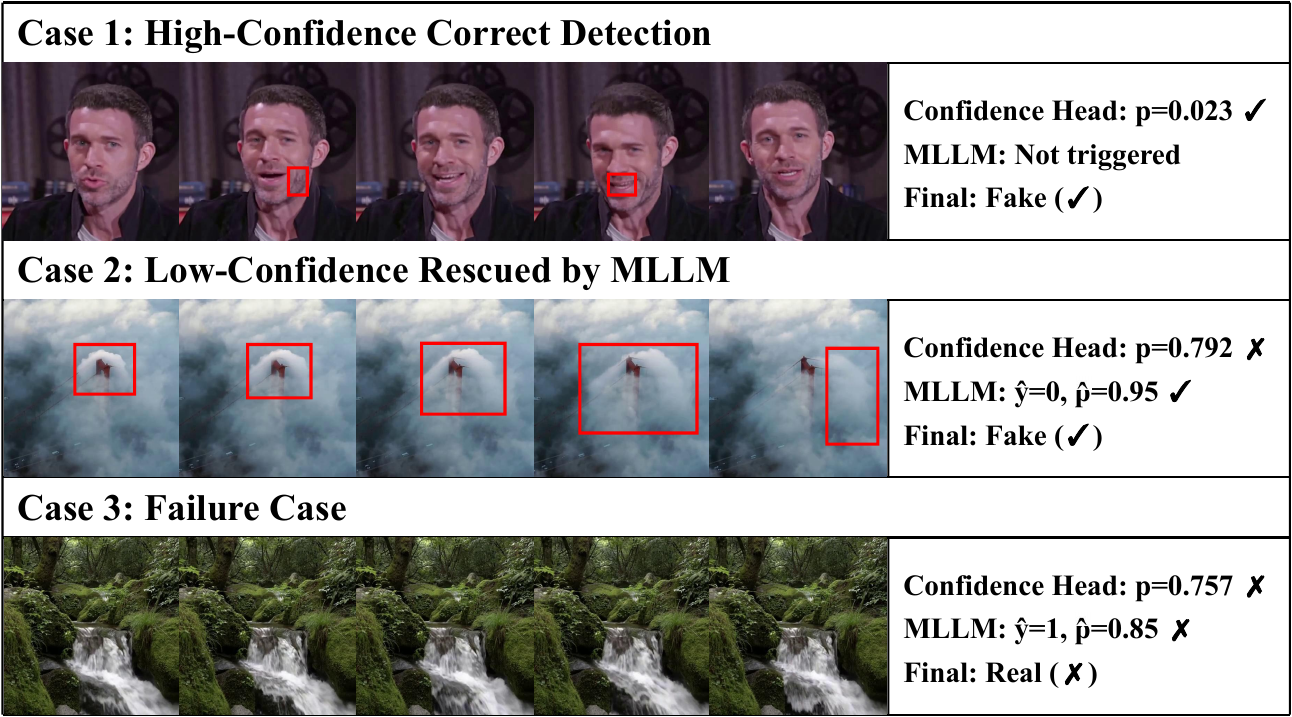}
   \caption{Case study of three fake video samples demonstrating different confidence scenarios and prediction outcomes.\vspace{-0.5em}}
   \label{fig:case_study}
   \vspace{-0.3em}
\end{figure}

To intuitively demonstrate how the confidence head of contrastive learning and MLLM influence the results under our confidence-gated detection mechanism, we analyze three representative fake video samples from the test set, as illustrated in Figure~\ref{fig:case_study}. 

\textbf{Case 1} demonstrates high-confidence detection where the contrastive network produces a strong confidence for fake classification ($p=0.023$), correctly identifying the fake video without MLLM intervention.
The evident facial artifacts (highlighted in red box) are effectively captured at the texture level.
\textbf{Case 2} illustrates the effectiveness of our confidence-gated mechanism.
The contrastive network's ambiguous prediction ($p=0.792$) triggers MLLM reasoning.
The MLLM identifies a semantic error where the suspension bridge is misinterpreted as a chimney by the generation model, successfully correcting the initial misjudgment with high confidence ($\hat{y}=0$, $\hat{p}=0.95$).
\textbf{Case 3} is a failure case where both modules misclassify the fake video.
The contrastive network fails to detect texture-level anomalies ($p=0.757$), while the simple valley stream scene lacks obvious semantic inconsistencies, leading the MLLM to also predict it as real ($\hat{y}=1$, $\hat{p}=0.85$).
This indicates that our method requires further improvement in capturing subtle artifacts within simple natural scenes.

More experiments on hyperparameter tuning, robustness, and MLLM comparisons can be found in the supplementary material.
\vspace{-0.5em}

\section{Conclusion}
\label{sec:conclusion}

In this work, we introduce \textbf{CoCoVideo-26K}, a high-quality AIGC video detection benchmark featuring contrastive real--fake pairs from diverse \textit{commercial} systems, enabling fine-grained texture analysis under semantic alignment.

We propose \textbf{CoCoDetect}, integrating contrastive learning with confidence-gated MLLM reasoning to capture texture-level and semantic-level inconsistencies.
Extensive experiments demonstrate that CoCoDetect consistently outperforms state-of-the-art methods, validating its robustness and generalization.

In the future, we plan to expand CoCoVideo with larger-scale multimodal data such as audio and richer semantic annotations, further facilitating comprehensive research on trustworthy AIGC content detection.
\section*{Acknowledgments}
Ming Zeng, Huidong Feng, Xinqi Cai, Yinglin Zheng, Yuxin Lin were partially supported by the National Natural Science Foundation of China (Grant No. 62072382) and the Yango Charitable Foundation.
{
    \small
    \bibliographystyle{ieeenat_fullname}
    \bibliography{main}
}

\clearpage
\setcounter{page}{1}

% Reset section counter
\setcounter{section}{0}
% Set section counter as A,B,C,D...
\renewcommand{\thesection}{\Alph{section}}

\maketitlesupplementary

% --------------------------------------------------------------
\section{Dataset Construction Details}

As mentioned in Section~\ref{datasetConstructionPipeline}, the overall framework of the dataset construction pipeline is shown in Figure~\ref{fig:construction}, and we elaborate on each of the four key stages in detail below.

\textbf{Stage I: Data acquisition and selection strategy.}
All real videos in our dataset are sourced from OpenVid-1M.
While deepfake research typically focuses on talking-head and human-interaction content, we deliberately expand the coverage to include food preparation, plant growth, cultural architecture, and natural landscapes to enhance generalizability across diverse AIGC scenarios.
Each selected video retains its textual description, which is directly reused as the generation prompt, and its first-frame image.
To ensure reliable and temporally aligned references, we only include videos with at least 5s of continuous footage within a single shot.
Category composition statistics are reported in Table~\ref{tab:distribution}.
Note that four models (Kling, Pika, Vidu, Vivago) were added during the second dataset expansion, resulting in slight differences in their category distributions compared to other models.

\textbf{Stage II: Two-round quality filtering.}
To ensure alignment and quality between generated outputs and real references, we apply two rounds of rigorous filtering.
The first round operates at the video level, removing clips that appear unrealistic or lack temporal coherence, including animation-like videos, pseudo-motion over static images, abrupt shot transitions, severe flicker, geometric distortions, and videos shorter than 5s.
The second round evaluates both the first-frame image and textual description.
For images, we discard samples with black first frames, heavy blur, or missing main subjects.
For text, we remove garbled or non-English descriptions to ensure consistent parsing, and filter out prompts exceeding platform length limits.
After these two rounds, all samples meet the first-frame and text requirements, ensuring semantic and visual alignment for subsequent generation.

\textbf{Stage III: Paired generation using commercial models.}
The filtered data are organized into batches of 1,000 videos according to specified category ratios.
For each real video, its first-frame image and textual description are used as input to generate the corresponding fake video via commercial video generation models.
Two generation modes are adopted: platform-based (through official web consoles) and API-based (via official interfaces).
When these two modes use different model versions (e.g., Jimeng via platform and Seedance via API), we treat them as distinct generation methods.
The generated video duration is fixed to 5 seconds for all models.
Resolution is matched to the original video when possible; otherwise, a higher resolution is used to maintain visual quality and comparability.
Across 13 commercial models, this stage produces a strictly matched fake counterpart for each real sample, yielding real--fake pairs with strong semantic and visual alignment.

\textbf{Stage IV: Post-processing normalization.}
We apply post-processing to all generated pairs and record detailed metadata for each video.
Videos with black borders or resolution mismatches are cropped and normalized.
However, when models automatically extend content near borders (e.g., Veo3 automatically expands a $720\times720$ source video to $1080\times720$), we retain these extended regions rather than cropping them back to preserve each model's native generation characteristics.
For each video pair, we provide detailed metadata including video name, caption (content description), camera motion, frame count, FPS, and duration.
The final dataset comprises 13,000 real--fake pairs (26,000 videos in total), forming CoCoVideo for subsequent training and validation.

% --------------------------------------------------------------
\section{Dataset Visual Examples}

\begin{figure*}[t]
  \centering
  \hspace{-2.5em}
   \includegraphics[width=0.93\linewidth]{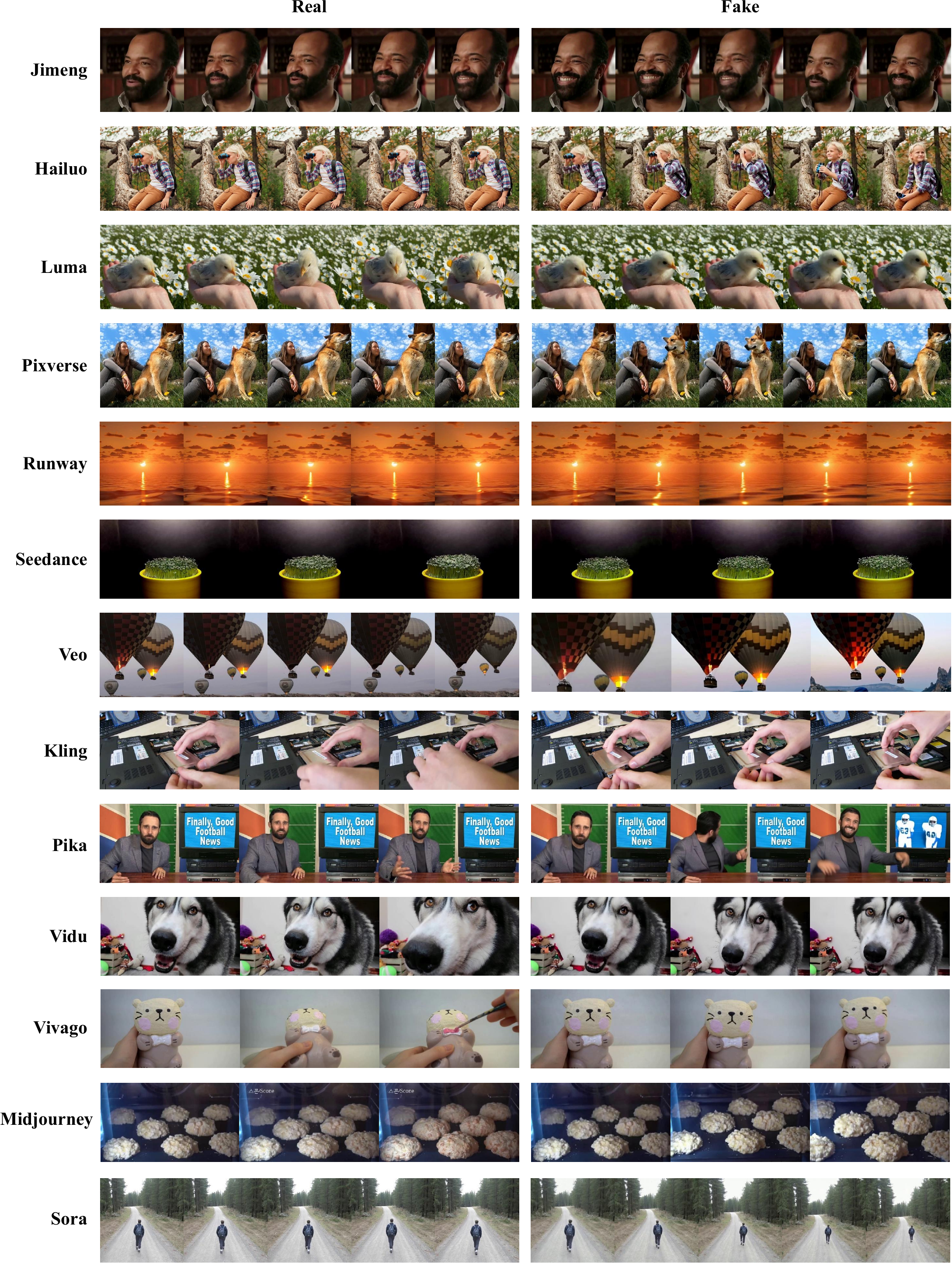}
   % \vspace{-0.5em}
   \caption{Visual examples of real--fake video pairs across different generation models.}
   \label{fig:visual_examples}
\end{figure*}

Figure~\ref{fig:visual_examples} presents visual examples from our dataset.
We select real--fake pairs across different categories for each generation model.
For videos with 1:1 aspect ratio, we display five uniformly sampled frames at 1-second intervals; for videos with 3:2 aspect ratio, we display three frames at 1s, 3s, and 5s.
The examples illustrate that generated videos are visually similar to their real counterparts and exhibit consistent temporal dynamics.

\begin{figure*}[t]
  \centering
   \includegraphics[width=1.0\linewidth]{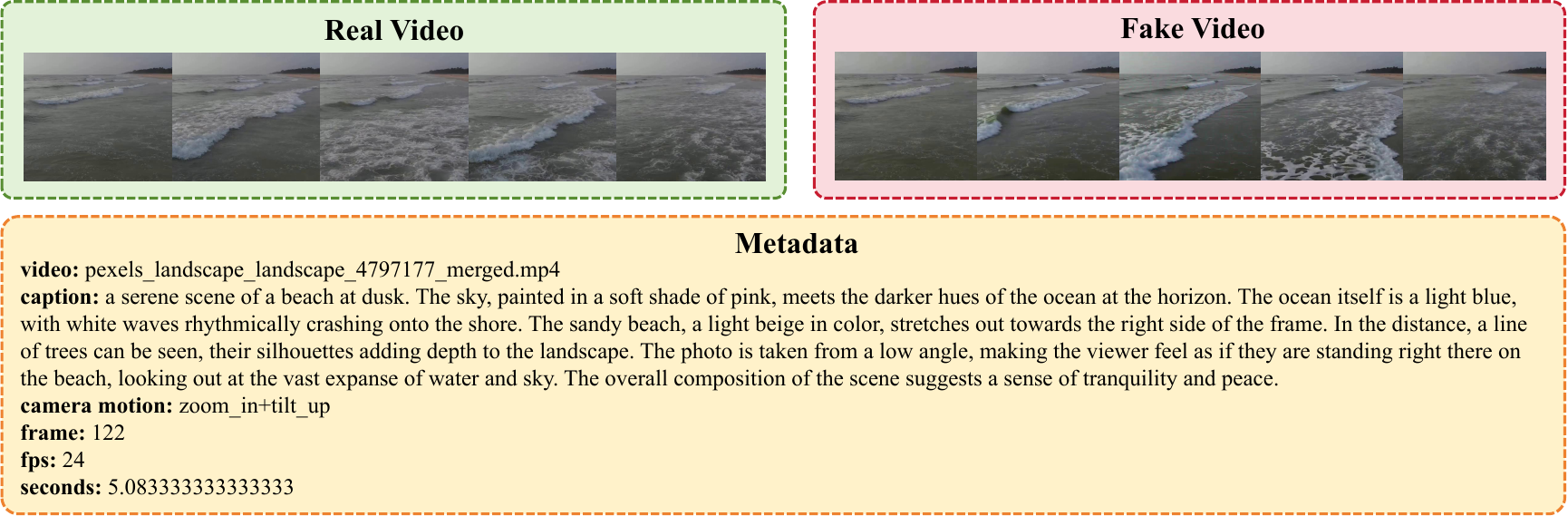}
   % \vspace{-0.5em}
   \caption{Example of the multimodal annotations provided for each video pair in our dataset.}
   \label{fig:metadata_example}
   % \vspace{0.5em}
\end{figure*}

Figure~\ref{fig:metadata_example} shows an example of the multimodal metadata provided for each video pair.
This example illustrates the annotations including video name, caption, camera motion, frame count, FPS, and duration, demonstrating the richness of metadata in CoCoVideo.

\begin{table*}[t]
  \centering
  \small
  \caption{Distribution of video category proportions (\%).}
  \label{tab:distribution}
  \renewcommand{\arraystretch}{1.1}
  \begin{tabularx}{0.9\textwidth}{@{}Y|YYYYYYY@{}}
    \toprule
    Model     & \makecell{Talking\\Head} & \makecell{Person\\Interaction} & \makecell{Cultural\\Architecture} & \makecell{Natural\\Landscape} & \makecell{Food\\Preparation} & \makecell{Plant\\Growth} & Others \\
    \midrule
      Jimeng     & 56.3 & 5.4 & 3.9 & 21.7 & 1.8 & 6.8 & 4.1 \\
      Hailuo     & 55.4 & 4.7 & 5.3 & 26.6 & 1.6 & 3.1 & 3.3 \\
      Luma       & 57.9 & 5.9 & 4.8 & 22.6 & 1.0 & 4.7 & 3.1 \\
      Pixverse   & 56.3 & 4.4 & 4.5 & 25.6 & 1.2 & 4.0 & 4.0 \\
      Runway     & 80.5 & 1.7 & 1.2 & 11.8 & 0.8 & 1.7 & 2.3 \\
      Seedance   & 56.3 & 5.7 & 6.2 & 24.0 & 1.1 & 3.1 & 3.6 \\
      Veo        & 51.2 & 4.0 & 5.9 & 28.4 & 1.1 & 5.7 & 3.7 \\
      Kling      & 28.0 & 57.9 & 1.2 & 3.5 & 1.6 & 4.0 & 3.8 \\
      Pika       & 30.2 & 49.7 & 0.6 & 3.6 & 2.4 & 2.6 & 10.9 \\
      Vidu       & 31.6 & 55.6 & 1.1 & 2.2 & 2.3 & 2.8 & 4.4 \\
      Vivago     & 33.3 & 48.5 & 1.4 & 2.1 & 1.3 & 2.5 & 10.9 \\
      Midjourney & 54.7 & 4.5 & 3.8 & 23.7 & 2.3 & 6.8 & 4.2 \\
      Sora       & 55.0 & 4.8 & 6.0 & 24.7 & 2.1 & 4.1 & 3.3 \\
    \bottomrule
  \end{tabularx}
  \vspace{-0.8em}
\end{table*}

% --------------------------------------------------------------
\section{More Experimental Setup Details}
\label{sec:experimentalSetupDetails}

\subsection{Parameters Details}
We train the model on a single NVIDIA A6000 GPU for 30 epochs, taking approximately 10 hours.
The input video resolution is set to $W = H = 224$ pixels with temporal dimension $T = 16$ frames, and a batch size of 8 video pairs is used.
During training, we apply data augmentation operations including: random horizontal flipping with probability 0.5 and color jittering (brightness=0.1, contrast=0.1, saturation=0.1, hue=0.05).
We use the AdamW optimizer with an initial learning rate of $1 \times 10^{-4}$ and weight decay of $1 \times 10^{-4}$, with a cosine annealing learning rate schedule over 30 epochs.

For the model-specific parameters, the classification loss weight is set to $\alpha = 0.65$, the paired contrastive loss uses a margin parameter of $m = 1.0$, and the confidence-gated inference mechanism employs a threshold of $\tau = 0.9$.

\subsection{MLLM Configuration}

\begin{figure*}[t]
  \centering
   \includegraphics[width=1.0\linewidth]{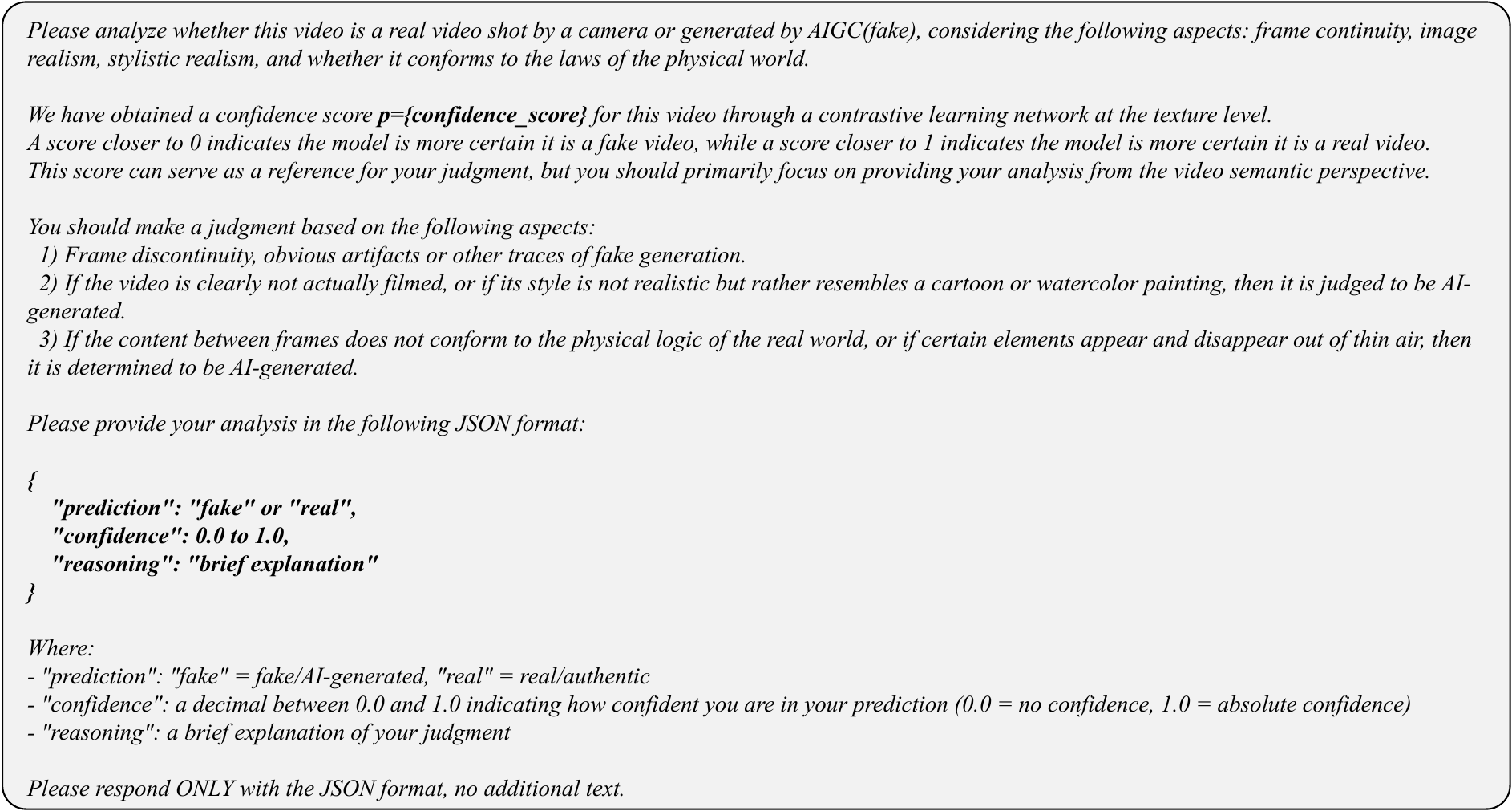}
   % \vspace{-0.3em}
   \caption{The text prompt provided to the MLLM for video authenticity reasoning.}
   \label{fig:prompt}
   \vspace{-0.8em}
\end{figure*}

We employ LLaVA-NeXT-Video-7B as the semantic reasoning model for analyzing uncertain samples routed by the confidence-gated mechanism.
The model is configured with maximum output tokens of 512 and temperature of 0.5, using the default frame sampling rate.
As shown in Figure~\ref{fig:prompt}, a task-specific prompt is used to guide the MLLM's reasoning process, which is jointly processed with the sampled video frames to generate the final authenticity prediction.

%-------------------------------------------------------------------------
\section{Additional Experiments}
\subsection{Hyperparameter Sensitivity Analysis}

\noindent\textbf{Loss Function Weights.}
We analyze the sensitivity of CoCoDetect to the loss function hyperparameters $\alpha$ and margin $m$ by fixing one parameter and varying the other.
To isolate the impact of the contrastive learning component, we conduct these experiments without MLLM integration.
Table~\ref{tab:loss_weights} compares the performance across different configurations in terms of Acc, F1, Recall, and AUC.
The results show that $\alpha=0.65$ and $m=1.0$ achieve the best Acc, F1, and AUC scores.
While some configurations yield higher recall, they sacrifice precision, leading to lower F1-Score and accuracy, confirming that our hyperparameter choice provides the optimal balance.

\begin{table}[t]
\centering
% \vspace{2.0em}
\caption{Performance under different loss weight configurations (\%). \textbf{Bold} indicates the best performance.}
\renewcommand{\arraystretch}{1.1}
\label{tab:loss_weights}
\begin{tabular*}{0.95\linewidth}{@{\hspace{2mm}}@{\extracolsep{\fill}} cc|cccc @{\hspace{2mm}}} 
\toprule
$\alpha$ & $m$ & Acc & F1 & Recall & AUC \\

\midrule
\multicolumn{6}{c}{\textit{Varying $\alpha$ (fixed $m=1.0$)}} \\
\midrule
0.5 & 1.0 & 87.79 & 88.05 & 89.85 & 94.41 \\
0.8 & 1.0 & 88.69 & 88.70 & 88.72 & 95.01 \\
0.9 & 1.0 & 87.08 & 87.00 & 86.46 & 93.62 \\
\midrule
\multicolumn{6}{c}{\textit{Varying $m$ (fixed $\alpha=0.65$)}} \\
\midrule
0.65 & 0.5 & 86.08 & 85.52 & 82.26 & 93.38 \\
0.65 & 1.5 & 83.38 & 84.97 & \textbf{93.90} & 93.18 \\
\midrule
\multicolumn{6}{c}{\textit{Final configuration}} \\ 
\midrule
\textbf{0.65} & \textbf{1.0} & \textbf{88.92} & \textbf{88.79} & 87.74 & \textbf{95.46} \\
\bottomrule
\end{tabular*}
\vspace{-0.5em}
\end{table}

\noindent\textbf{Confidence Threshold Selection.}
The confidence threshold $\tau$ controls the trade-off between detection accuracy and sample coverage in the contrastive learning module.
Table~\ref{tab:threshold_analysis} compares the performance under different threshold values ranging from 0.6 to 0.9.
As $\tau$ increases, the accuracy improves but the sample coverage decreases.
To balance these two metrics, we compute an $F_\beta$-like score:
\begin{equation}
F_\beta = \frac{(1+\beta^2) \cdot \text{Cov} \cdot \text{Acc}}{\beta^2 \cdot \text{Cov} + \text{Acc}}
\end{equation}
where \textit{Acc} denotes accuracy and \textit{Cov} denotes coverage rate.
We set $\beta=2$ to emphasize accuracy, which is prioritized in this detection task.
The results show that $\tau=0.9$ achieves the highest $F_2$ score, providing the optimal balance.

\begin{table}[t]
\centering
\vspace{1.0em}
\caption{Performance comparison under different confidence thresholds (\%). \textbf{Bold} indicates our choice with the best F$_2$-Score.}
\renewcommand{\arraystretch}{1.1}
\label{tab:threshold_analysis}
\begin{tabular*}{0.95\linewidth}{@{\hspace{2mm}}@{\extracolsep{\fill}} c|ccc @{\hspace{2mm}}} 
\toprule
$\tau$ & Accuracy  & Coverage  & F$_2$-Score  \\
\midrule
0.6 & 89.03 & 99.62 & 90.97 \\
0.7 & 89.26 & 99.08 & 91.06 \\
0.8 & 89.49 & 98.59 & 91.18 \\
\midrule
\textbf{0.9} & \textbf{89.77} & \textbf{97.51} & \textbf{91.22} \\
\bottomrule
\end{tabular*}
\end{table}

\subsection{Backbone Comparison}

To validate the effectiveness of our backbone choice, we conduct comparative experiments with different architectures.
Table~\ref{tab:backbone_comparison} compares the performance of ResNet-18 with other commonly used backbones including ResNeXt, VideoMAE, and MViT-v2.
The results show that ResNet-18 achieves the best F1, Acc, and AUC scores.
Although MViT-v2 achieves higher recall, it results in lower F1, Acc, and AUC due to reduced precision.
We attribute this to the unique characteristics of our dataset, where real and fake videos share strong semantic correlations due to the paired generation process.
More complex architectures with higher capacity tend to overfit to subtle training-specific patterns, leading to reduced generalization compared to the simpler ResNet-18 structure.
This finding suggests that lightweight models are more suitable for detecting AI-generated videos with high semantic similarity to their real counterparts.

\begin{table}[ht]
\centering
\small
% \vspace{0.1em}
\caption{Performance comparison of different backbone architectures (\%). \textbf{Bold} indicates the best performance.}
\renewcommand{\arraystretch}{1.1}
\label{tab:backbone_comparison}
\resizebox{1.0\linewidth}{!}{
\begin{tabular}{l|cccc}
\toprule
Backbone & Acc & F1 & Recall & AUC  \\
\midrule
ResNeXt & 80.49 & 80.65 & 81.33 & 88.16 \\
VideoMAE & 79.64 & 77.98 & 72.10 & 89.26 \\
MViT-v2 & 86.21 & 86.95 & \textbf{91.90} & 94.28 \\
\midrule
\textbf{ResNet-18 (Ours)} & \textbf{88.92} & \textbf{88.79} & 87.74 & \textbf{95.46} \\
\bottomrule
\end{tabular}
}
\end{table}

\subsection{Robustness Analysis}

To evaluate the robustness of CoCoDetect under realistic perturbations, we apply various distortions to the test videos and measure the performance degradation.
We consider six common perturbation types with the following configurations: (1) CRF compression with quality factor CRF=28 to simulate video transmission compression; (2) grayscale conversion to remove all color information; (3) Gaussian noise with standard deviation $\sigma=0.1$ added to normalized pixel values; (4) Gaussian blur with kernel size $5\times5$ and $\sigma=2.0$ to simulate defocus; (5) geometric transformation including random rotation within $\pm15°$ and scaling by factor $0.9\sim1.1$; and (6) local occlusion with 20\% of the frame area randomly masked.
Table~\ref{tab:robustness} presents the performance of CoCoDetect under these perturbations in terms of Acc, F1-Score, and AUC.
The results show that CoCoDetect maintains relatively stable performance under geometric transformation and local occlusion.
MLLM helps correct some misclassified samples through semantic reasoning.
However, Gaussian noise and blur cause more significant performance drops, as these perturbations lead to high-confidence incorrect predictions by the contrastive learning module, preventing samples from being routed to MLLM for correction.

\begin{table}[ht]
\centering
\caption{Robustness evaluation of CoCoDetect under different perturbations. All metrics are in percentage (\%). \textbf{Bold} indicates the best performance.}
\renewcommand{\arraystretch}{1.1}
\label{tab:robustness}
\resizebox{1.0\linewidth}{!}{
\begin{tabular}{l|ccc}
\toprule
Perturbation & Acc & F1 & AUC \\
\midrule
CRF Compression (28) & 74.82 & 77.61 & 85.43 \\
Grayscale & 76.13 & 73.51 & 84.19  \\
Gaussian Noise ($\sigma=0.1$) & 59.49 & 62.16 & 63.01  \\
Gaussian Blur ($\sigma=2.0$) & 66.87 & 68.16 & 69.01 \\
Geometric Transform ($\pm15°$) & 84.92 & 85.60 & 92.85 \\
Local Occlusion (20\%) & 83.41 & 84.69 & 92.82  \\  
\midrule
\textbf{None (Original)} & \textbf{90.69} & \textbf{90.62} & \textbf{95.93} \\
\bottomrule
\end{tabular}
}
\end{table}

\subsection{MLLM Selection Study}

To select the most suitable MLLM for semantic reasoning, we conduct a comprehensive comparison across three key dimensions: response speed, reasoning accuracy, and output format compliance (the ability to consistently generate structured outputs adhering to the required JSON format).
We use 500 videos from the GVD dataset as our evaluation benchmark for two reasons: (1) their short duration requires fewer frames to be transmitted to the MLLM, enabling faster response times; (2) all videos in GVD are AI-generated with obvious semantic inconsistencies that violate real-world physics, making it effective for validating whether MLLMs can identify physical implausibilities.

We compare four representative 7B-parameter models: LLaVA-NeXT-Video-7B, Qwen2.5-VL-7B-Instruct, LLaVA-1.5-7B-hf, and DeepSeek-VL2.
Table~\ref{tab:mllm_comparison} summarizes the results.
LLaVA-NeXT-Video-7B achieves the highest reasoning accuracy with near-perfect format compliance, demonstrating the best balance for our task despite slightly longer response time than LLaVA-1.5-7B-hf.
While LLaVA-1.5-7B-hf has the fastest response time, its significantly lower accuracy makes it unsuitable for this task.
Therefore, we adopt LLaVA-NeXT-Video-7B in our framework.
Note that due to computational constraints, we restrict our evaluation to 7B models; larger models may yield further improvements in reasoning accuracy, though at the cost of increased inference time and resource requirements.

\begin{table}[ht]
\centering
\caption{Comparison of different MLLMs on the GVD dataset. \textit{T} denotes Response Time (seconds), \textit{Acc} denotes Accuracy (\%), and \textit{F.C.} denotes Format Compliance (\%). \textbf{Bold} indicates the best performance.}
\renewcommand{\arraystretch}{1.1}
\label{tab:mllm_comparison}
\begin{tabular*}{0.95\linewidth}{@{\hspace{2mm}}@{\extracolsep{\fill}} l|ccc @{\hspace{2mm}}} 
\toprule
Model & T & Acc & F.C. \\
\midrule
LLaVA-1.5-7B-hf & \textbf{2.19} & 44.00 & \textbf{100.00} \\
DeepSeek-VL2 & 3.28 & 79.60 & 95.40 \\
Qwen2.5-VL-7B-Instruct & 3.91 & 92.40 & 97.60 \\
\midrule
\textbf{LLaVA-NeXT-Video-7B} & 3.72 & \textbf{95.38} & 99.60 \\
\bottomrule
\end{tabular*}
\end{table}

\begin{figure}[ht]
  \centering
  % \vspace{-0.3em}
   \includegraphics[width=1.0\linewidth]{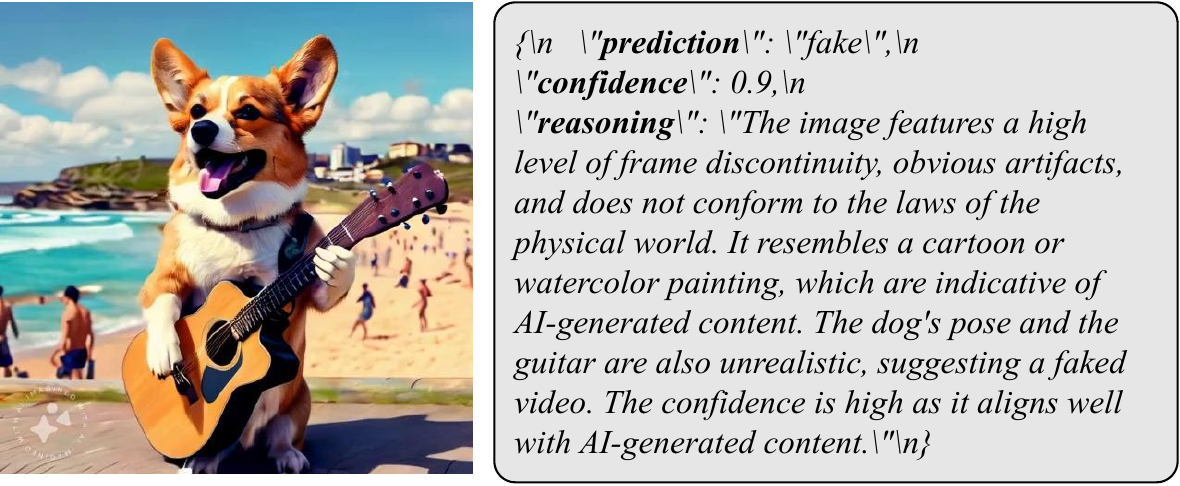}
   \caption{Example of correctly formatted MLLM output with proper JSON structure.}
   \vspace{-0.5em}
   \label{fig:mllm_output}
\end{figure}

Figure~\ref{fig:mllm_output} illustrates a correctly formatted MLLM response from LLaVA-NeXT-Video-7B.
Common format compliance failures include: (1) adding extraneous conversational text (e.g., \textit{"Sure, here is my JSON output..."}); (2) omitting curly braces around JSON content; and (3) missing quotation marks around field names or values.
These issues highlight the importance of format compliance in our evaluation.

\section{Copyright and Ethical Considerations}
\label{sec:ethicalAndCopyright}

We prioritize copyright compliance throughout the construction of CoCoVideo-26K to ensure legal and transparent data sourcing.
For the real-video component, all samples are sourced exclusively from the OpenVid-1M dataset, which is released under a CC-BY-4.0 license that allows redistribution and adaptation with proper attribution.
We follow the attribution requirements specified by the license in all downstream usage.
For the synthetic-video component, we employ only official commercial model platforms or their authorized APIs, which operate under standard usage policies that permit generated content to be used for academic research purposes.
It aims to align the generation process with model providers' terms of service, thereby reducing the risk of unauthorized use or potential copyright infringement.

Beyond legal compliance, we carefully consider the ethical implications of constructing and releasing a deepfake detection benchmark.
Our primary motivation is to advance defensive technologies against malicious AI-generated content, thereby contributing to a safer digital media ecosystem.
We deliberately avoid including sensitive categories to minimize potential misuse while maintaining research value.
The dataset is made available and released exclusively for academic research purposes.
By transparently documenting our data sources, generation methods, and ethical safeguards, we aim to promote responsible research on AI-generated media that balances scientific advancement with societal responsibility.

\end{document}